\documentclass[10pt,twocolumn,letterpaper]{article}

\usepackage{multirow}
\usepackage{tabularx}
\usepackage[table]{xcolor}
\usepackage{arydshln}
\usepackage{wrapfig}
\usepackage{makecell}
\usepackage{pifont}
\usepackage{afterpage}
\usepackage{footnote}

\usepackage{tikz}
\usepackage{soul}
\usepackage{adjustbox}

\usepackage[sectionbib]{chapterbib}

\newcommand{\cmark}{\ding{51}}%
\newcommand{\xmark}{\ding{55}}%

\newcommand{\coloruline}[2]{\setulcolor{#1}\setul{}{1pt}\ul{#2}\setulcolor{black}}
\newcommand{\coloredcircle}[1]{
    \begin{adjustbox}{valign=c}
        \tikz\draw[#1, fill=#1,draw=black] (0,0) circle (1.1ex);
    \end{adjustbox}
}

\makeatletter
\newcommand*{\textoverline}[1]{$\overline{\hbox{#1}}\m@th$}
\makeatother

\newcommand{\customhl}[2]{%
    \begingroup
    \sethlcolor{#1!50} 
    \hl{#2}%
    \endgroup
}

\usepackage[pagenumbers]{cvpr} 

\usepackage[pagebackref,breaklinks,colorlinks,citecolor=cvprblue]{hyperref}
\usepackage{circledsteps} 

\def\paperID{17547}
\def\confName{CVPR}
\def\confYear{2024}

\newcommand{\red}[1]{{\color{red}{#1}}}

\newcommand{\datasetName}{Stellar}

\newcommand{\webpage}{https://stellar-gen-ai.github.io}

\def\eg{e.g.,}

\definecolor{cvprblue}{rgb}{0.21,0.49,0.74}

\title{Stellar: Systematic Evaluation of Human-Centric\\Personalized Text-to-Image Methods}

\author{
    \textbf{
    Panos Achlioptas\thanks{All authors made very significant and distinct contributions. Email Correspondence: stellar@steelperlot.com}\hspace{20pt}
    Alexandros Benetatos\footnotemark[1]\hspace{20pt}
    Iordanis Fostiropoulos\footnotemark[1]\hspace{20pt}
    Dimitris Skourtis\footnotemark[1]}\\
    [.2cm]
    Steel Perlot Management LLC\\
    Los Angeles, California, USA
}

\begin{document}
\maketitle

\begin{abstract}
In this work, we systematically study the problem of personalized text-to-image generation, where the output image is expected to portray information about specific human subjects. E.g., generating images of oneself appearing at imaginative places, interacting with various items, or engaging in fictional activities. To this end, we focus on text-to-image systems that input a single image of an individual to ground the generation process along with text describing the desired visual context. Our first contribution is to fill the literature gap by curating high-quality, appropriate data for this task. Namely, we introduce a standardized dataset (Stellar) that contains personalized prompts coupled with images of individuals that is an order of magnitude larger than existing relevant datasets and where rich semantic ground-truth annotations are readily available. Having established Stellar to promote cross-systems fine-grained comparisons further, we introduce a rigorous ensemble of specialized metrics that highlight and disentangle fundamental properties such systems should obey. Besides being intuitive, our new metrics correlate significantly more strongly with human judgment than currently used metrics on this task. Last but not least, drawing inspiration from the recent works of ELITE~\cite{elite} and SDXL~\cite{sdxl}, we derive a simple yet efficient, personalized text-to-image baseline that does not require test-time fine-tuning for each subject and which sets quantitatively and in human trials a new SoTA. 
For more information, please visit our project's website:~\url{\webpage}.
\end{abstract}    
\section{Introduction}
\label{sec:intro}
There is no doubt that the recent progress in terms of developing and publicly sharing multimodal deep learning models trained with internet scale data, comprised of billions of textual tokens and images (e.g., \cite{clip_metric,stable_diffusion,sdxl}), have given rise to an unprecedented explosion of novel solutions for many long-standing problems at the intersection of computer vision and natural language~\cite{yin2023survey,chatgpt_multimodal}. Applications ranging from creating de novo fictional images~\cite{dall-e,Imagen,dalle-2} to editing them via text~\cite{brooks2022instructpix2pix,zhang2022sine}, or similarly manipulating higher-dimensional visual representations (e.g., video~\cite{godiva,han2022tell,singer2022makeavideo,menapace2023plotting} or 3D objects~\cite{CLIP-Forge,achlioptas23_shapetalk,dreambooth3d,zhang2023text2nerf}); have vastly pushed the envelope for computer-aided \textit{creativity} and further democratized this field given the accessibility of their linguistic interfaces. 

Among the plethora of text-to-modality methods, we focus in this work on text-to-image (T2I) models that can \textit{personalize} their output by considering input text that specifies imaginative renditions of specific \textit{human individuals}. Such personalized models have recently started to emerge~\cite{elite,textual_inversion,dreambooth,shi2023instantbooth} as a natural and powerful extension of de novo image generators~\cite{stable_diffusion,sdxl}, inspiring radically novel applications. These applications include personalized story-telling via character animation of specific individuals, aiding people to attend otherwise inaccessible real-world events virtually, or simply lifting the burden to capture photographs physically~\cite{animation_genai,characters_genai,dreams_snap}. Despite these encouraging prospects, two noticeable limitations currently hinder future developments in this field.

First, there is a need for carefully curated, \textit{standardized} data, explicitly targeting  personalized T2I. Current efforts make it easy to find vast quantities of \textit{objective} captioning data for images (e.g., ~\cite{LAION-5B,OpenImages}); however, curating images of individuals performing \textit{fictional} actions accompanied by textual descriptions is practically impossible given the open-ended and imaginary nature of this task~\cite{dreambooth}. Even if one is to lift the constraint of coupling the descriptions of images to specific individuals~\cite{dreambooth,parti,kumari2022customdiffusion}, there is yet no common ground for building large-scale collections of textual prompts in \textit{natural} language that can be deemed applicable for generating i) interesting fictional depictions of humans, being ii) diverse, and iii), carrying semantically meaningful meta-annotations that can be used to evaluate the output quality of different personalized T2I approaches.

Second, more rigorous evaluation protocols are needed for comparing methods and tangibly measuring progress on personalized T2I tasks. Despite that, many recent and indispensable metrics for \textit{de novo} text-to-image systems exist~\cite{clip_metric,LAION-5B,hessel2021clipscore}; such metrics are typically blind to fundamental aspects that affect the overall performance and quality of personalized, subject-driven generators. For instance, the ability of a model to maintain the distinctive facial characteristics of the input subject, i.e., its `identity' at its output, while essential for a successful generation, is typically left unchecked. To make matters worse, recent works (e.g., \cite{sdxl,PickaPic}) have shown that widely used metrics, such as CLIP~\cite{clip_metric} or FID~\cite{FID}, correlate poorly with human preferences, even for subject-agnostic de novo image generations.

In this work, we aim to address the above shortcomings and set robust foundations to assist future developments in personalized T2I generation. In summary, we introduce \Circled{\textbf{1}} a large-scale multimodal dataset, \textbf{\datasetName}, with prompts of imaginary human-centric depictions, grounding publicly available images of celebrities~\cite{CelebAMask-HQ}. \Circled{\textbf{2}} an extensive evaluation framework comprised of intuitive metrics capturing complementary and fundamental aspects determining the efficacy of modern personalized T2I models. We use these metrics to make a thorough evaluation of existing SoTA personalized neural methods and, lastly, we develop \Circled{\textbf{3}} a simple, yet efficient, T2I baseline which sets a new SoTA both per algorithmic evaluations and in human judgment assessed via extensive human evaluation trials.

\begin{table}[t]
    \centering
    \caption{Comparison of {\datasetName} with existing datasets focusing on personalized, or de novo (Parti~\cite{parti}), image generations. {\datasetName} has more prompts and identities than any existing dataset, focuses explicitly on human-centric depictions, and provides rich meta-annotations for \textit{both} the paired prompts \& image-based identities.}
    \label{table:stellar_data_comparisons}
    {\scriptsize
        \begin{tabularx}{\linewidth}{>{\centering}m{3cm}|cccc}
            \toprule
            Dataset & {\makecell{\#\\Prompts}} & {\makecell{\#\\Subjects}} & {\makecell{Human-\\Centric}} & {\makecell{Meta-\\Annot.}} \\
            \midrule
            DreamBooth~\cite{dreambooth} & $35$ & $40$ & \xmark & \xmark \\
            CustomConcept101~\cite{kumari2022customdiffusion} &$617$ & $101$ & \xmark & \xmark \\
            Parti~\cite{parti} & $1.6k$ & N/A & \xmark & \cmark \\
            \midrule
            Stellar (Ours) & $\mathbf{20k}$ & $\mathbf{400}$ & \cmark & \cmark \\
            \bottomrule
        \end{tabularx}
    }
\end{table}
\begin{table*}[ht]
    \centering
    \caption{\textbf{Representative prompts of Stellar-$\mathcal{H}$ along with template-based examples of Stellar-$\mathcal{T}$}. The Template column showcases the underlying production rule followed to generate the Stellar-$\mathcal{T}$ example of each row. The right-most column captures the \textit{object-centric} categories we assign to each row example of Stellar-$\mathcal{T}$. Different colors indicate object-based categories named by their corresponding part in the \textit{Template} column, \eg \protect\coloredcircle{orange} represents the \textit{fruit-related category}.}

    {
        \rowcolors{2}{gray!10}{}
        \begin{tabularx}{\textwidth}{p{142pt}|p{132pt}|p{140pt}|p{33pt}}
            \textbf{Stellar-\texorpdfstring{$\mathcal{H}$}} & \textbf{Stellar-\texorpdfstring{$\mathcal{T}$}} & \textbf{Template} & \textbf{Categ.}\\

            \midrule

            juggling oranges on a rooftop & juggling \coloruline{orange}{apples} in the \coloruline{pink}{airport} & juggling \customhl{orange}{[fruit]} \customhl{pink}{[building loc.]} & \multirow{1}{*}{\coloredcircle{orange} \coloredcircle{pink}}\\

            riding a skateboard in outer-space & riding a \coloruline{teal}{skateboard} near \coloruline{magenta}{Saturn} & riding \customhl{teal}{[vehicle]} \customhl{magenta}{[space loc.]} & \multirow{1}{*}{\coloredcircle{teal} \coloredcircle{magenta}}\\

            wrestling an octopus on a pirate boat & \multirow{2}{*}{fighting a \coloruline{purple}{shark} on a \coloruline{teal}{boat}} & \multirow{2}{*}{fighting \customhl{purple}{[animal]} on \customhl{teal}{[vehicle]}} & \multirow{2}{*}{\coloredcircle{purple} \coloredcircle{teal}}\\

            as an astronaut walking to a rocket & as an \coloruline{olive}{astronaut} near a \coloruline{gray}{spaceship} & as \customhl{olive}{[uniformed]} near \customhl{gray}{[space-object]} & \multirow{1}{*}{\coloredcircle{olive} \coloredcircle{gray}}\\

            \multirow{2}{*}{playing the trombone in a castle} & \multirow{2}{*}{playing the \coloruline{cyan}{saxophone} in a \coloruline{pink}{castle}} & playing \customhl{cyan}{[musical instrument]} \customhl{pink}{[building loc.]} & \multirow{2}{*}{\coloredcircle{cyan} \coloredcircle{pink}}\\

            riding an elephant through the jungles of Thailand & \multirow{2}{*}{riding a \coloruline{purple}{tiger} in the \coloruline{brown}{jungle}} & \multirow{2}{*}{riding \customhl{purple}{[animal]} \customhl{brown}{[nature loc.]}} & \multirow{2}{*}{\coloredcircle{purple} \coloredcircle{brown}}\\

            eating a croissant at a cafe in Paris & eating a \coloruline{blue}{souvlaki} in \coloruline{red}{Athens} & eating \customhl{blue}{[food]} \customhl{red}{[city]} & \multirow{1}{*}{\coloredcircle{blue} \coloredcircle{red}}\\

            \bottomrule
        \end{tabularx}
    }
    \label{tab:prompts_examples}
\end{table*}

\section{Related Works}
\label{sec:related_works}

\paragraph{Grounded Visio-Linguistic Datasets.}
    There is a vast literature and accompanying datasets concerning objective multimodal captioning of images~\cite{coco_chen2015,referit,LAION-5B,plummer2015flickr30k,conceptual-captions,google_ref,VG_Krishna_2017,OpenImages}, 3D objects and scenes~\cite{achlioptas2019shapeglot,text2mesh,3DScanEnts,achlioptas2020referit_3d,zhenyu2019scanrefer,SNARE}, or even more subjective bi-modal data concerning e.g., visually grounded emotions~\cite{achlioptas2021artemis,artemis_V2,achlioptas2022affection}, or dynamic actions predicated on visual stimuli~\cite{park2020visualcomet}. In stark contrast, datasets containing textual \textit{prompts} capable of rendering personalized subject-driven generations aided by T2I AI, are a scarce commodity. DiffusionDB~\cite{diffusion_db} is a large-scale dataset that includes user prompts that are not personalized and typically contain multiple complex modifiers (\eg~famous painters, photographic techniques, etc.), resulting in obscure, not natural language. Recent personalized datasets (e.g., ~\cite{dreambooth,parti,kumari2022customdiffusion}) are not focused on human-centric actions, cover few subjects, and/or lack rich meta-annotations accompanying the paired prompts/images. These shortcomings are addressed in \datasetName~(refer to Table~\ref{table:stellar_data_comparisons} for a short contrastive summary).

\paragraph{Metrics for T2I Generations.}
    Several metrics in the literature aim to evaluate different aspects of de novo T2I systems. For example, metrics such as CLIP$_{T}$~\cite{elite} and CLIP-Score~\cite{hessel2021clipscore} attempt to evaluate holistically the semantic alignment between a sentence and a given image. More recent approaches, such as PickScore~\cite{PickaPic} and HPS~\cite{wu2023better}, aim at algorithmically evaluating the alignment of textually-grounded generations and user-preference based on ground-truth preferential user data. Orthogonally to these metrics, unimodal (vision-based) metrics, such as the Aesthetic-Score~\cite{LAION-5B}, try to quantify the aesthetic quality of a given image. Besides the poor correlation shown between CLIP-based metrics and human-preference~\cite{sdxl,PickaPic}, and the fact that \textit{subjective} quantities such as an image's aesthetic value are intrinsically hard to compute~\cite{achlioptas2021artemis}; all the above metrics are \textit{not} designed to take into account the additional image portraying the subject that is driving the generation. I.e., they are effectively oblivious to the personalized angle we are focusing on.

\paragraph{Personalized T2I Systems.} 
    Capitalizing on the remarkable capabilities and modularity shown recently by T2I models~\cite{stable_diffusion,sdxl,dalle,Imagen}, powerful extensions personalizing their output generations have emerged. Starting with the pioneering works of Gal \textit{et. al}~\cite{textual_inversion} and Ruiz \textit{et. al}~\cite{dreambooth}, personalized networks with admirable capabilities have started to appear. One important limitation of these two initial works was their lack of efficiency since they required a separate model optimization (in the form of finetuning) for each input subject. More recently, works like ELITE~\cite{elite}, or similar ones(~\cite{gal2023encoderbased,jia2023taming}), have lifted this bottleneck by learning more general and robust mappings between visual and textual embeddings capable of generalizing simultaneously to many individual inputs. Similar to these later works, our introduced network (StellarNet) also avoids a per-subject-based optimization and, per extensive quantitative and qualitative analysis, a SoTA alternative.
\section{Personalized Text-to-Image Data}\label{sec:dataset}

The goal of the Stellar (\textit{\underline{S}ys\underline{t}ematic \underline{E}va\underline{l}uation} of Persona\underline{l}ized Im\underline{a}ge\underline{r}y)  dataset is to provide a large-scale standardized evaluation dataset for human-based personalized T2I generation tasks. Stellar is comprised of \textbf{20,000} carefully curated prompts (selected from a \textit{pool of \textbf{187k} candidate prompts}) describing imaginative situations and actions for rendering human-centric fictional images. Moreover, Stellar's prompts are paired to \textbf{400} \textit{unique} human identities, selected from CelebAMask-HQ~\cite{celeba, CelebAMask-HQ}; and all paired items come with rich meta-annotations promoting a rigorous evaluation of personalized T2I systems, as described in the following paragraphs.

\paragraph{Stellar Prompts.}    
    For the textual part of Stellar, our goal is to offer a large set of natural prompts that describe imaginative renderings of humans involved in many activities, real-world object interactions, portrayed at distinct locations, etc. In doing so, we only enforce two crucial constraints. First, we restrict our prompts from including commands that \textit{explicitly} ask to alter the physical characteristics of an individual, such as their age, sex, eye color, or other facial characteristics that, in large, remain constant over long periods. Besides the many existing methods and corresponding data (e.g.,~\cite{stylegan,fader_networks,FACEGAN,ShenL16b,CAFE-GAN,celeba,CelebAMask-HQ} \textit{specializing} in editing/describing such characteristics - the derived invariance to such attributes eases the evaluation between the input/output subject (by assessing if such characteristics remained unchanged). Our second restriction concerns requests that could otherwise intentionally control the \textit{fine-grained} appearance of a human in terms of clothes, eyewear, footwear, etc. (e.g., \textit{wearing a red dress}). We believe that T2I generators handling apparel in a try-on fashion deserve unique attention and future work. 
    
    With the above setup established, we first curate Stellar-$\mathcal{H}$(uman) comprised by 10,000 prompts given by English-speaking expert annotators\footnote{We point the reader to Supp. for a detailed explanation of \textit{all} AMT experiments and safeguarding protocols promoting high-quality annotations.}, recruited via Amazon Mechanical Turk (AMT). Upon reading a detailed set of instructions and inspecting examples of good vs. bad prompts, the annotators were free to type natural prompts they would use to create images of \textit{themselves} assuming access to an imaginative (and powerful) personalized T2I model. The resulting corpus is remarkably diverse and imaginative, containing 6.4$k$ distinct words in 8-word-long sentences on average. It contains $\sim$1.5$k$ verbs and 4.5$k$ nouns, and the average length-normalized pairwise cosine distance of its sentences computed in an ST5-based~\cite{sentence_t5} embedding-space, is \textbf{1.58} larger ($p$ $<$$.001$) than the average distance among captions of COCO~\cite{coco_chen2015}. Typical examples are shown in Table~\ref{tab:prompts_examples} and an extensive analysis in our Supplementary Material~\cite{stellar_supp}.
    
    \begin{figure*}[ht]
        \begin{minipage}{0.49\textwidth}
            \includegraphics[width=\linewidth]{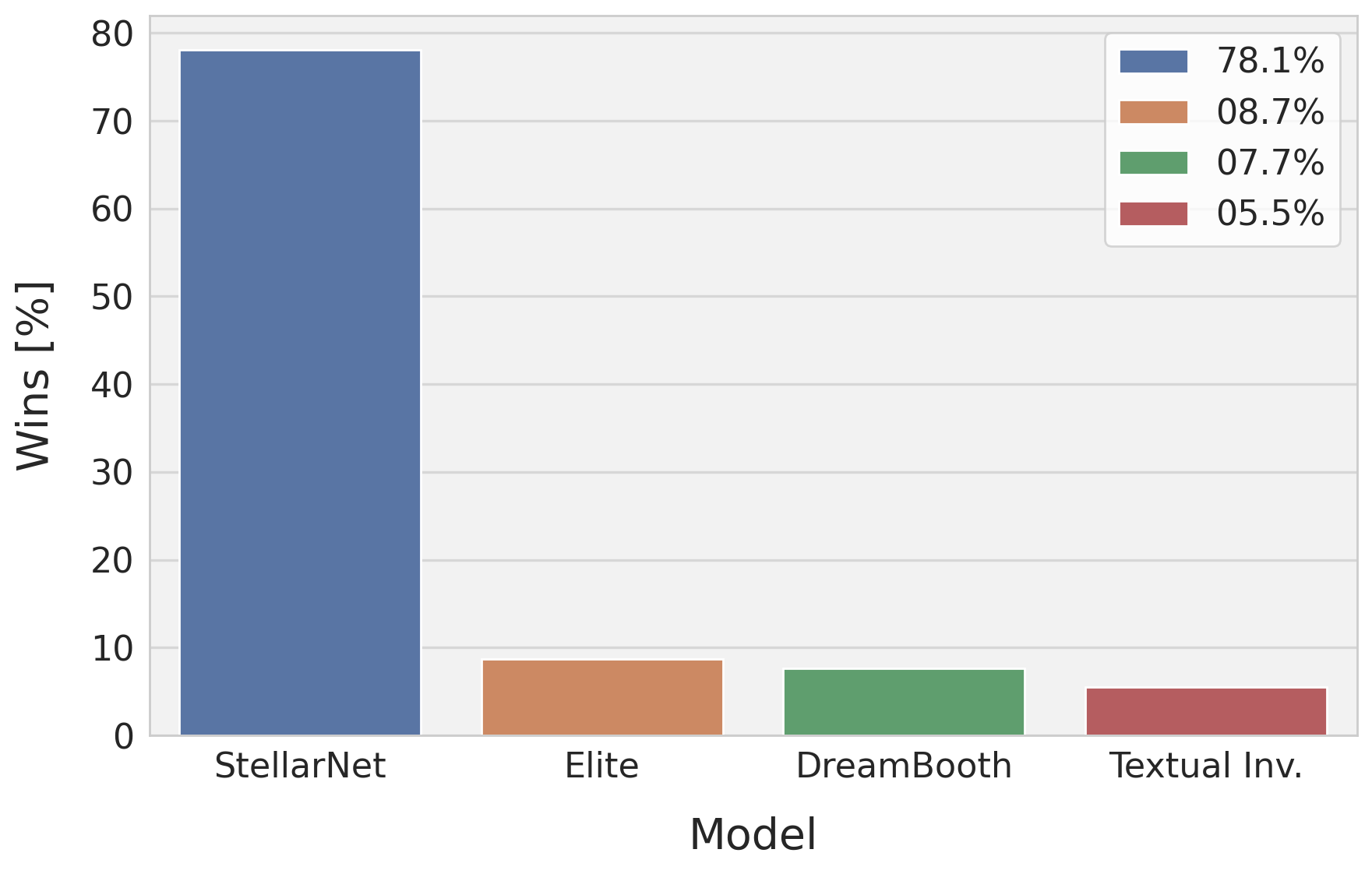}
        \end{minipage}
        \begin{minipage}{0.49\textwidth}
            \centering
            \begin{tabular}{lcc|c}
                \toprule
                \multirow{2}{*}{Metric} & \multicolumn{3}{c}{Human Pref. \textit{Kendall}-$\tau$} \\
                \cmidrule{2-4}
                 & Obj & Rel  & Overall \\
                \midrule
                CLIP$_T$ & 0.130 & 0.089 & 0.106\\
                HPSv2 & 0.067 & 0.004 & 0.224\\
                PickScore & 0.139 & 0.110 & 0.246\\
                DreamSim & 0.031 & 0.132 & 0.270\\
                CLIP$_I$ & 0.049 & 0.011 & 0.304\\
                HPSv1 & 0.094 & 0.103 & 0.304\\
                ImageReward & 0.040 & 0.060 & 0.320\\
                Aesth. & 0.049 & 0.146 & 0.359\\
                \midrule
                GOA & \textbf{0.175} & 0.110 & 0.149\\
                RFS & 0.121 & \textbf{0.163} & 0.167\\
                APS & 0.013 & 0.018 & 0.389\\
                SIS & 0.112 & 0.011 & 0.435\\
                IPS & 0.094 & 0.018 & \textbf{0.455}\\
                \bottomrule
            \end{tabular}
        \end{minipage}

        \caption{\textbf{\textit{Left:}} User preferences between StellarNet and existing personalized approaches. StellarNet's output is preferred by a very large margin (\textit{78\%} of all trials). \textbf{\textit{Right: }} Kendall's ($\tau$) correlation among existing and our introduced metrics. Our metrics correlate significantly better with key aspects of personalized generations and human-preference. }
        \label{fig:amt-study}
    \end{figure*}

    While the diversity mentioned above and the open-endedness of Stellar-$\mathcal{H}$ make it a realistic dataset for an intrinsically imagination-based task, they can also place certain limitations in its usage when analyzing personalized T2I systems. Namely, given the ongoing performance-differential of close vs. open-vocabulary discriminative methods~\cite{wu2023open}, Stellar-$\mathcal{H}$'s variance and complexity make its evaluation harder with classical methods. To compensate for this limitation and further standardize Stellar's prompts, we \textit{augment} Stellar with 10,000 additional prompts generated in a semi-automatic and controlled way. Specifically, we bootstrap Stellar-$\mathcal{H}$ by fitting to it abstract \textit{templates} (e.g., [person] eating [food] at [place]) that can be used to derive the majority ($>$75\%) of its prompts.

    Critically, such templates enable us to \textit{flexibly} generate new sentences that follow Stellar-$\mathcal{H}$'s lexical structure and semantics, but where we can control the sets of their constituent parts. For, we we can restrict nouns associated with [food] to specific (e.g., detectable~\cite{VG_Krishna_2017}) food items. In practice, we use a mixture of \textit{frequently} appearing objects and object-to-object relations by co-analyzing Stellar-$\mathcal{H}$ with well-established object-detection datasets (COCO~\cite{coco_chen2015}, Objects365~\cite{Shao2019Objects365AL}, OpenImages~\cite{OpenImages}); as well as using recommendations from ChatGPT~\cite{chatGPT} regarding typical examples from broad categories extracted from the aforementioned resources. The exhaustive generation of all underlying production rules gives rise to $177k$ candidate prompts. We sub-sample those to 10$k$ final prompts, constituting Stellar-$\mathcal{T}$(emplate), by simultaneously keeping a high-variance (using farthest-point-sampling~\cite{fps}) and semantic-proximity to Stellar-$\mathcal{H}$ per ST5's embedding space.

    Notably, the above process achieves the implicit meta-annotation of our prompts. For instance, in Stellar-$\mathcal{T}$, $350$ unique detectable objects exist, with an average of $1.5$ `detectable' objects per prompt. Moreover, its prompts cover $45$ distinct predicates and $\sim$$1k$ unique relational triplets of the form \texttt{<human}-\texttt{predicate}-\texttt{object>}. Last but not least, we manually curate discrete categories (``themes") characterizing all prompts (see Table~\ref{tab:prompts_examples}). The semantics captured by this meta-data will be indispensable in evaluating T2I approaches in Section \ref{sec:experimental_results}. 

\paragraph{Stellar Subject Images.} 
    Equipped with the above prompt-based part of Stellar, we further couple these prompts with high-resolution human-portraying images existing in CelebAMask-HQ's~\cite{CelebAMask-HQ}. In doing so, we sampled 400 human subjects, ensuring that the resulting image-collection accompanying Stellar satisfies the following desiderata. First, it is \textit{unbiased} to the sex and age of its subjects (per Celeb's binary labels - see original images of Fig.~\ref{fig:seed-ablation} which span these categories). Second, it covers subjects for which more than one distinct image exists. Finally, all its subjects and images are annotated with ground-truth physical attributes and segmentation masks separating each human from the background. Those latter masks are helpful for evaluation purposes and for training modern T2I generators where background-free grounding images typically result in qualitatively superior generations~\cite{elite}. Finally, we note that we combine each subject with 50 prompts chosen uniformly at random, without repetition, to partition our 20$k$-large text collection fully.

\section{Metrics}\label{sec:metrics}

\begin{table*}[!ht!]
    \centering
    \caption{\textbf{Quantitative evaluation of StellarNet against popular SoTA alternatives using existing and introduced metrics.}}
    \label{table:quant_stellar_vs_sota}
    
    \begin{tabular*}{\textwidth}{l|ccccc}
         & \multicolumn{4}{c}{\textit{Models}}  \\
        \textbf{Metrics} & \textbf{DreamBooth}~\cite{dreambooth} & \textbf{ELITE}$^{*}$~\cite{elite} & \textbf{Text. Inv.}~\cite{textual_inversion} & \textbf{StellarNet} (Ours) & \textit{Type}\\

        \midrule

        Aesth. ($\uparrow$)~\cite{LAION-5B}  & 5.316 & 5.095 & 5.263 & \textbf{5.713} & \multirow{3}{*}{Img-to-Img}\\
        CLIP$_I$ ($\uparrow$)~\cite{clip_metric} & 0.301 & 0.387 & 0.477 & \textbf{0.531} & \\
        DreamSim ($\downarrow$)~\cite{fu2023dreamsim} & 0.789 & 0.691 & 0.608 & \textbf{0.566} & \\

        \midrule

        CLIP$_T$ ($\uparrow$)~\cite{clip_metric} & \textbf{0.413} & 0.365 & 0.311 & 0.382 & \multirow{5}{*}{Text-to-Img}\\
        HPSv1 ($\uparrow$)~\cite{wu2023better} & 0.197 & 0.192 & 0.188 & \textbf{0.202} & \\
        HPSv2 ($\uparrow$)~\cite{wu2023human} & 0.276 & 0.272 & 0.267 & \textbf{0.278} & \\
        ImageReward ($\uparrow$)~\cite{xu2023imagereward} & -0.049 & -0.510 & -0.998 & \textbf{0.320} & \\
        PickScore ($\uparrow$)~\cite{PickaPic} & 0.212 & 0.205 & 0.198 & \textbf{0.212} & \\

        \midrule

        GoA ($\uparrow$) & 0.302 & 0.260 & 0.229 & \textbf{0.305} & \multirow{5}{*}{\makecell{Personalized \\ (Ours)}}\\
        RFS ($\uparrow$) & 0.103 & 0.106 & 0.082  & \textbf{0.134} & \\
        APS ($\uparrow$) & 0.317 & 0.490 & 0.510 & \textbf{0.693} & \\
        SIS ($\uparrow$) & 0.232 & 0.355 & 0.262 & \textbf{0.577} & \\
        IPS ($\uparrow$) & 0.252 & 0.383 & 0.287 & \textbf{0.637} & \\

        \midrule

        Human Preference ($\uparrow$)  & 0.077 & 0.087 & 0.055  & \textbf{0.781}  &\\

        \bottomrule
    \end{tabular*}
\end{table*}

To the best of our knowledge, there are not yet adopted metrics specializing in generative personalized imagery.
Instead, existing evaluation methods for T2I models (personalized or not) use a combination of metrics originally designed for \textit{unconditional} image generation, (\eg~FID~\cite{FID} or IS~\cite{salimans2016improved}), and bimodal metrics such as CLIP~\cite{clip_metric} (or extensions~\cite{hessel2021clipscore,xu2023imagereward,wu2023better}) which are optimized for \textit{de novo} T2I generations that are not subject-driven.

To make matters worse, humans exhibit a heightened critical awareness of nuanced distortions or inaccuracies in representations of human faces~\cite{bruce1986understanding,bentin1996electrophysiological}. For instance, while minor imperfections in the rendering of inanimate objects like bananas might go unnoticed, even a slight deviation in the portrayal of human facial features can be met with a strong, adverse reaction. This disparity in evaluative standards underscores the need for personalization systems focused on human-driven generations to adhere to a higher level of precision and accuracy, as human faces are not just scrutinized more closely but also evoke stronger emotional responses~\cite{disfigured_faces} upon perceived inaccuracies~\cite{uncanny_valley}.

In what follows, we assume a method that accepts as input the image of a human subject ($I$), along with a prompt ($P$), and is tasked to produce an output image ($O$). 

\subsection{Evaluating identity-centric qualities} 
    We introduce three novel metrics aimed at quantifying the similarity between the input subject's `identity', and the generated output image; (i) a coarse similarity metric assessing identity preservation (ii) a fine-grained metric focused on facial attribute preservation, and (iii) a metric orthogonal to the first two, designed to evaluate the sensitivity and robustness of a T2I method in maintaining the subject's identity consistently under novel input renderings. Prior to introducing these metrics, we address a key challenge inherent to `naive' image-to-image-based metric evaluations, setting the stage for a comprehensive discussion of our proposed methodologies.

    \paragraph{Metric Exploitation.}
        In evaluating image-to-image correspondences, it is crucial to recognize that pairwise metrics directly measuring image similarity (e.g., CLIP$_I$), will score perfectly a method that merely replicates the input image to the output, disregarding the prompt. To counteract this, we adapt such metrics to \textit{soft}-penalize output generations that deviate significantly from the input prompt. 
        
        Specifically, we utilize CLIP$_T$~\cite{clip_metric}, to independently measure the similarity between the prompt and \textit{both} the input and output images. An output is penalized if it fails to improve on the input/prompt similarity by at least 2 standard deviations $\sigma$, calculated across the distribution of CLIP$_T$ scores on all Stellar test data. Given that $\sigma$ is calculated across a large collection, and our focus is on capturing pronounced overfitting modes rather than nuanced visual differences, the particular choice of text-to-image metric (\eg~CLIP$_T$, PickScore) is of limited significance. Thus, the binary penalty $\mathbb{I}_p$ is calculated as follows:

        \begin{equation}\label{eq:naive}
            \mathbb{I}_p = [\text{CLIP}_T(P, I) \times 2\sigma_{\text{CLIP}_T}]< \text{CLIP}_T(P, O)
        \end{equation}

    \paragraph{Identity Preservation Score.}
        \texttt{IPS} is designed to assess the facial resemblance between the input identity and the generated images in a rather coarse but specialized way. Our metric uses a face detector \cite{Deng2020CVPR} to isolate the identity's face in both input and generated images. It then employs a specialized face detection model to extract facial representation embeddings from the detected regions. \texttt{IPS} is calculated using the metric (cosine)similarity between the embeddings from the input image and output image whose face confidence is above a selected threshold $\alpha$.

        Two main considerations must be addressed when evaluating a method with \texttt{IPS}. Firstly, the metric becomes inapplicable if the output lacks a human subject. Secondly, multiple subjects in the output introduce ambiguity in identifying the input identity.

        To counter these issues, we assign zero scores to outputs without a detectable face. Additionally, for outputs with multiple faces, we compute their similarity against the input identity and use the maximum as the \texttt{IPS}. Lastly, we use the penalty, $\mathbb{I}_p$ (Eq. (\ref{eq:naive})), to calibrate the plausible exploitation of image-to-image metrics. Thus, if $f$ represents the facial feature extractor, our metric is formulated as:

        \begin{equation}
            \texttt{IPS} = \max f(I) \cdot \mathbb{I}_p [ f_{>\alpha}(O) ]
        \end{equation}

    \paragraph{Attribute Preservation Score.}
        \texttt{APS} focuses on assessing how well the generated images maintain specific fine-grained attributes of the identity in question, such as age, gender, and other invariant facial features (\eg~high cheekbones). Leveraging the annotations in Stellar images, we can evaluate these binary facial characteristics.

        Specifically, we train a series of linear classifiers on the face embeddings extracted from CelebAHQ, utilizing the corresponding ground-truth attribute annotations. These classifiers are employed to independently predict the finer-grained attributes in the generated images of each subject.

        To compute \texttt{APS}, considering the imbalanced distribution of the attribute labels we use the ROC-AUC score for each attribute and report their average. Additionally, we incorporate the penalization from Eq. (\ref{eq:naive}). Let $M$ denote the total number of attributes, with $p_m$ representing the classifiers for the $m$-th attribute. Further, let $\mathbf{f_{O}}$ be the outputs of the face-detector and $\mathbf{Y_{m}}$ the annotations of the $m$-th attribute. With this notation, the \texttt{APS} is formulated as:

        \begin{equation}\label{eq:attr}
            \texttt{APS} = \frac{1}{M}\sum_{m=1}^{M}{\text{AUC} \left[  \mathbb{I}_n \left[ p_{m}(\mathbf{f_{O}}), \mathbf{Y_{m}} ) \right] \right]}
        \end{equation}

    \paragraph{Stability of Identity Score.}
        \texttt{SIS} explores an orthogonal dimension to the rest of our identity-based metrics. Namely, it serves as a measure for determining the extent of a model's sensitivity to \textit{different} images of the \textit{same} individual; further promoting models where the subject's identity is consistently well-captured irrespective of the input's image irrelevant \textit{variations} (e.g., lighting conditions, subject's pose).         
        To achieve this goal, \texttt{SIS} necessitates having access to multiple images of the human subject (a condition met in Stellar's dataset by design); and is our only evaluation metric with such a more demanding requirement.

        If $S$ represents the total number of the subject's images available (all coupled with the same prompt), $I_m$ is the $m$-th input image, and $O_m$ the corresponding $m$-th output generation, \texttt{SIS} is computed as follows:

        \begin{equation}
            \texttt{SIS} = \frac{1}{S} \sum_{m\in S} \min_{k\in S\setminus\{m\}}\,\texttt{IPS}(I_k, O_m)
        \end{equation}

        Notice that by forcing the similarity measurement between unaligned input/output pairs ($k\in S\setminus\{m\}$), we naturally dampen the potential of \texttt{IPS} to overfit to irrelevant for one's identity details, since such, are less likely to persist across different (input) images. Furthermore, to reduce the increased computational requirement of generating outputs for multiple input images per identity, a fast `approximation' of the above metric can be formulated using only a \textit{single} output image $O_m$, as:
        \begin{equation}
            \texttt{SIS}_{fast} = \min_{k\in S\setminus\{m\}}
            \,\texttt{IPS}(I_k, O_m)
        \end{equation}

\subsection{Object-centric Context Evaluation}
    Established bimodal metrics assessing text-image similarity (\eg~CLIP$_T$) often fall short in providing granular insights into their scoring rationale. For instance, given a low similarity, it is impossible to reason whether this stems from a failure to depict the prompt's objects, their requested interactions, the desired atmosphere or style, or combinations of the above. To overcome this limitation, we introduce specialized and interpretable metrics to evaluate two key aspects of the alignment between the output image and the input prompt: object representation faithfulness, and the fidelity of depicted relationships.

    \paragraph{Grounding Objects Accuracy.} \label{para:metrics:context:co-apperance}
        \texttt{GOA} aims to evaluate the visual faithfulness of grounding context objects referenced in the prompt at the generated image. In contrast to CLIP's holistic approach (involving the comparison between the entire image and the entire prompt), \texttt{GOA} is a \textit{localized} metric that allows us to decouple the evaluation of the subject's identity preservation from that of correctly grounding the referenced objects of the prompt.

        To compute \texttt{GOA}, we utilize an object detector \cite{OWL2} to infer localized visual objects (in the form of bounding boxes) for each object referenced in the prompt, and average their prediction confidences, $cf$. Relying on each object appearing at most once in the Stellar-$\mathcal{T}$ prompts, if multiple bounding boxes are found for a referenced object, we utilize the one with the highest confidence ignoring the others. If $Y_m$ represents the $m$-th referenced object in a single prompt out of $M$ in total, and $d$ is the object detector used that generates detection confidences, $cf$, then \texttt{GOA} is formulated as:

        \begin{equation}
            \texttt{GOA} = \frac{1}{M} \sum_{m=1}^{M} \max_{cf} d(Y_{m}, O)
        \end{equation}

    \paragraph{Relation Fidelity Score.}
        \texttt{RFS} aims to evaluate the success of representing the desired prompt-referenced object interactions on the generated image. Considering the difficulty of even specialized Scene Graph Generation (SGG) models to understand visual relations, this metric introduces a valuable insight into the ability of the personalized model to faithfully depict the prompted relations.

        Our approach leverages an SGG model \cite{PENET} to extract relationship triplets in the form of \texttt{<subject} - \texttt{predicate} - \texttt{object>} from the generated images. First, given the set of ground truth relation labels from Stellar-$\mathcal{T}$, we map those to the vocabulary of the SGG model, following \cite{alex_sgg, SGGfromNLS}, to get the metric labels, $Y$. If $G$ is the generated graph from the SGG model, we only keep triplets with a human \texttt{subject}, and additionally, pick triplets from $G$ if their object matches the objects in $Y$. This way, for each filtered triplet, $k$, we have an associated annotated relation, $R_{k}$, from Stellar-$\mathcal{T}$, and a probability distribution vector, $\mathbf{D_k(}O\mathbf{)}$ from $G$. Letting the probability $P_k$ of the $R_{k}$ relation from $\mathbf{D_k(}O\mathbf{)}$, be denoted as $P_k(O, R_k)$, and averaging over all $N$ relations, we define \texttt{RFS} as:

        \begin{equation}
            \texttt{RFS} = \frac{1}{N} \sum_{k=1}^{N}{P_k(O, R_k)}
        \end{equation}

\section{StellarNet Baseline}\label{sec:method}

\begin{figure}[t]
    \centering
    \includegraphics[width=0.95\linewidth]{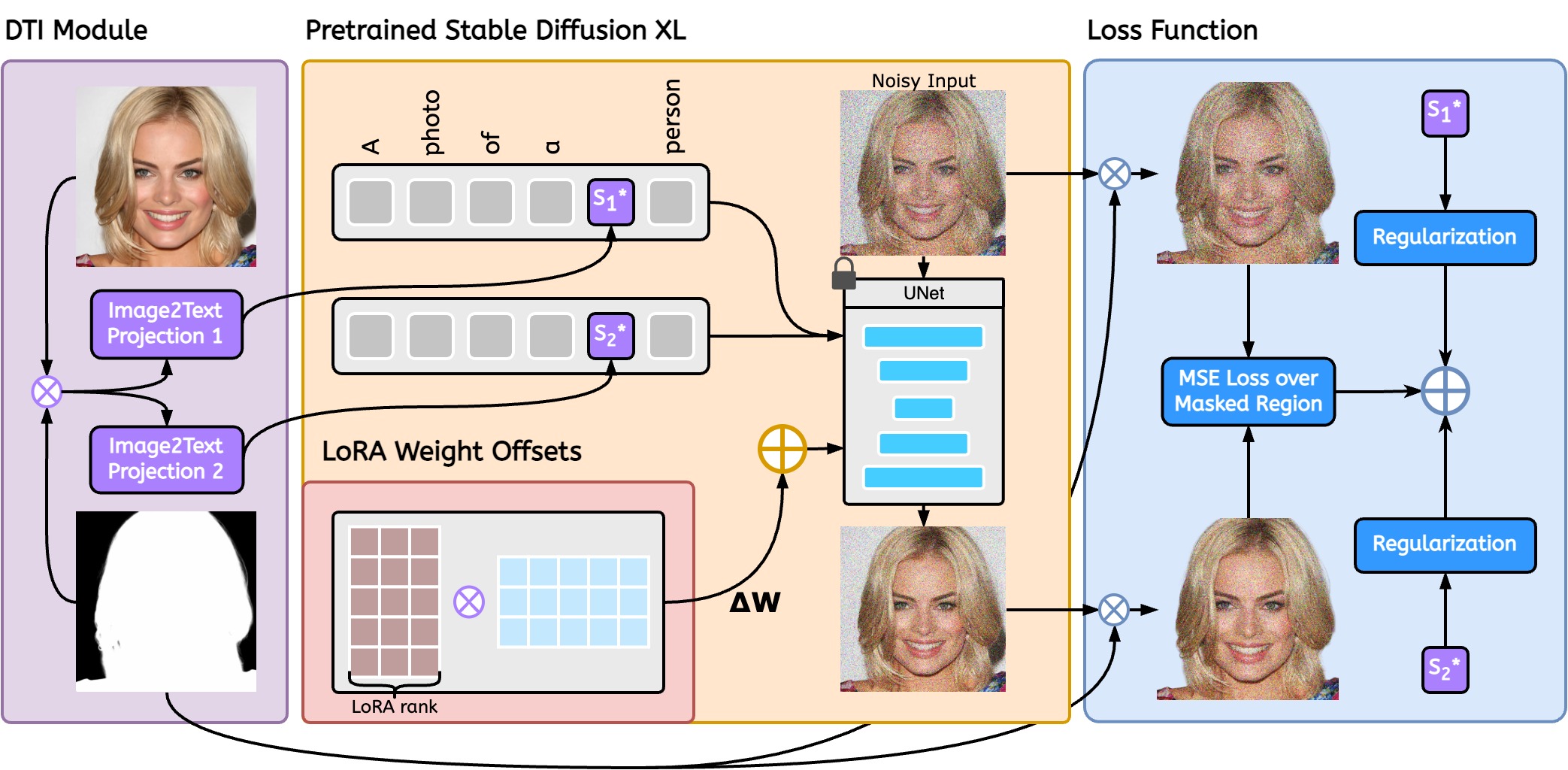}
    \caption{\textbf{StellarNet overview}. The Dynamic Textual Inversion (\texttt{DTI}) module inverts the foreground-masked identity image
    (found in CelebAMask-HQ~\cite{CelebAMask-HQ})    
    into textual embeddings, $S^*$ (\textbf{\textcolor{violet}{left}}). The $S^*$ augments the textual prompt passed to the pre-trained text-to-image model (SDXL~\cite{sdxl}) to guide the model into generating images with the given identity (\textbf{\textcolor{orange}{middle}}). Additionally, we finetune the UNet backbone of SDXL using LoRA weight-offsets~\cite{hu2022lora} for efficient and stable training (\textbf{\textcolor{purple}{middle bottom}}). We apply a \textit{masked} MSE loss during training over the input image \textit{and} the output generation (\textbf{\textcolor{blue}{right}}).}
    \label{fig:stellar_architecture}
\end{figure}

To analyze and evaluate human-centric personalized T2I methodologies more comprehensively, we introduce a new baseline model, drawing substantial inspiration from prior approaches. Our primary objective is \textit{not} to devise a new personalized T2I \textit{paradigm} but to meaningfully enrich our comparative assessment pool. Secondarily, we wish to highlight the potential of leveraging more powerful open-source, non-personalized T2I systems (\eg~SDXL~\cite{sdxl}) in combination with well-curated data (\eg~CelebAMask-HQ~\cite{CelebAMask-HQ}) to assist personalized generators.

Similar to existing research~\cite{textual_inversion, dreambooth}, our architecture relies on large pre-trained T2I diffusion models~\cite{stable_diffusion, sdxl} tapping on their excellent capacity for prompt-guided image generation. Following recent works \cite{elite,gal2023encoderbased}, originally inspired by GAN-Inversion techniques~\cite{gan_inversion}, we also aim to derive a latent representation for the provided subject image, $I$, in the pre-trained word-embedding space of a T2I model. Assuming learning a successful \textit{projection} of $I$ in this space, a personalized generator can then treat the subject as a word and leverage the compositionality of natural language to synthesize novel images involving the subject.

\paragraph{Dynamic Textual Inversion.} To achieve the above goal efficiently and overcome the need for per-subject optimization, we incorporate an encoder-based inversion technique (similar to~\cite{elite, gal2023encoderbased}). To this end, we use image encoders to acquire image embeddings for $I$ and map those to textual embeddings (denoted as $\mathbf{S^*}$) through shallow Multilayer Perceptrons (MLP). During training, we optimize $\mathbf{S^*}$,  comprised of a series of embeddings projected from different layers of the image encoder. Still, we use only the last layer’s embedding during inference to de-emphasize lower-abstraction details learned from the input image, enhancing the model’s focus on the subject’s salient characteristics. Additionally, to better guide the network to focus on the input subject, we use binary masks to separate it from the background of the images passed to the image encoders. We refer to this process and its underlying neural module as Dynamic Textual Inversion (\texttt{DTI}) (Fig. \ref{fig:stellar_architecture} - left).

\paragraph{SDXL Incorporation.} At the core of our architecture lies the SDXL~\cite{sdxl} model (base model, excluding the refiner), extended with a \texttt{DTI} module that augments the textual input with embeddings representing the personalization input. The \texttt{DTI} module utilizes two CLIP image encoders (CLIP ViT-L \& OpenCLIP ViT-bigG), aligned with the original SDXL's two CLIP text encoders. Additionally, we finetune the pre-trained UNet backbone of SDXL to more effectively interpret the new $\mathbf{S^*}$ embeddings. Efficient and stable training for the larger SDXL is achieved using Low-Rank Adaptation (LoRA)~\cite{hu2022lora} weight offsets (Fig. \ref{fig:stellar_architecture} - middle bottom). Finally, to avoid unnecessary model penalization due to irrelevant input details, during training, we also apply a \textit{masked} MSE loss over the input and output generation (Fig. \ref{fig:stellar_architecture} - right). More implementation details can be found in Sec. \red{3} - Supp.~\cite{stellar_supp}.

\begin{figure}[t!]
    \centering
    \includegraphics[width=\linewidth]{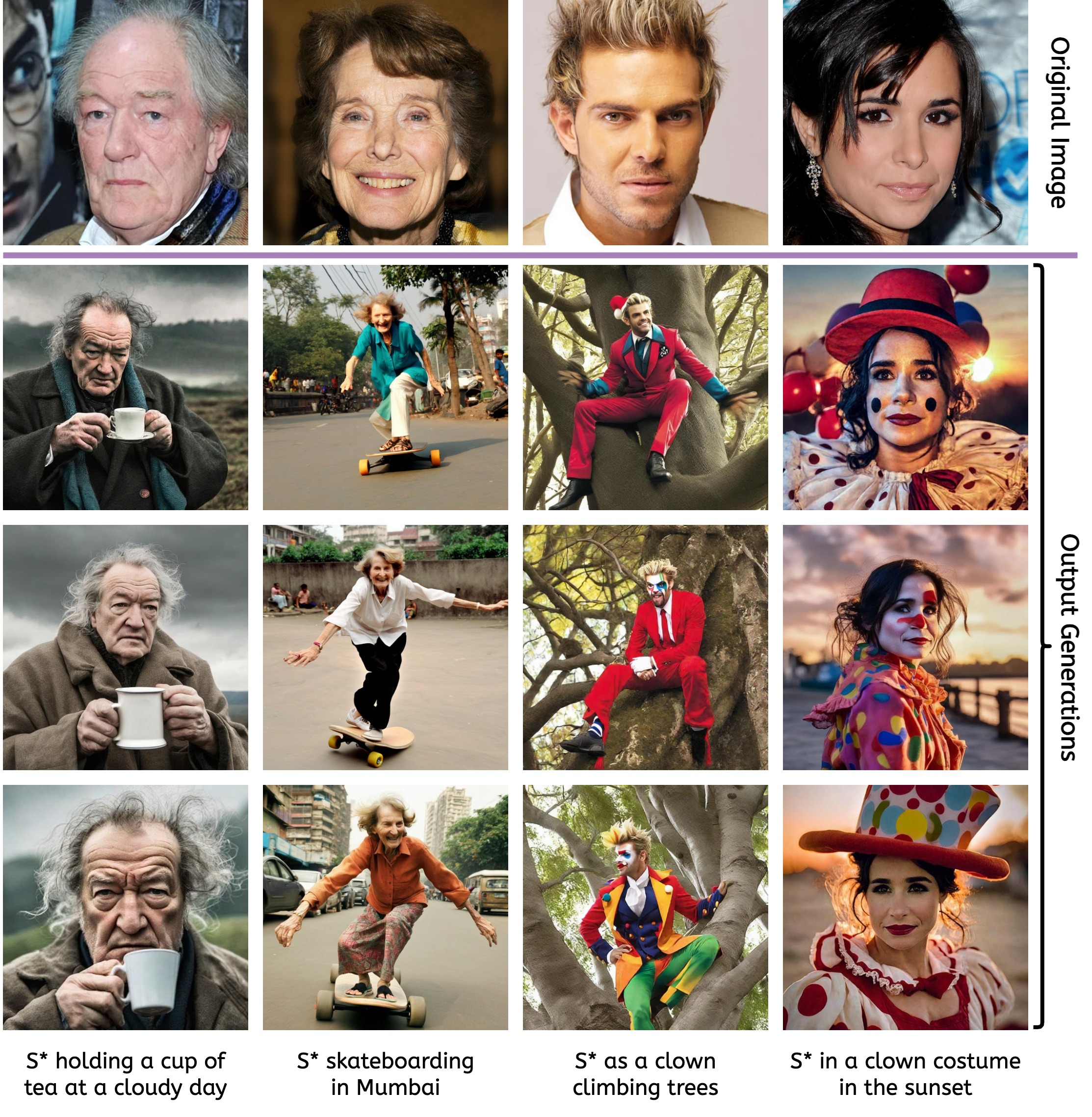}
    \caption{\textbf{StellarNet generations grounded on different noise-controlling random seeds.} StellarNet produces rich variations given a human subject
    (top row~\cite{CelebAMask-HQ}) and a fixed prompt (bottom of each column). Best viewed by zooming in on the digital version.}
    \label{fig:seed-ablation}
\end{figure}

\section{Experimental Results}
\label{sec:experimental_results}

In this section, we detail the experiments conducted to \Circled{\textbf{1}} validate the alignment of our Metrics (\cref{sec:metrics}) with human judgment, \Circled{\textbf{2}} contrast them with existing metrics, and \Circled{\textbf{3}} show the effectiveness of the StellarNet Baseline (\cref{sec:method}) compared to popular personalized generators. 
Our experimental analysis is partly possible due to the introduction of the Stellar Dataset that enables the usage of our fine-grained metrics over a large-scale and standardized benchmark.

In all the experiments we evaluate along StellarNet, DreamBooth (DB)~\cite{dreambooth}, Textual Inversion (TI)~\cite{textual_inversion}, and ELITE \cite{elite}. Specifically, we train StellarNet and ELITE on the \textit{official} training set of CelebAMask-HQ consisting of \textit{24k} images. We denote ELITE trained on this image dataset as ELITE$^*$. Note that human identities and their underlying images in Stellar (evaluation) Dataset do \textit{not} overlap with the training set of CelebAMask-HQ. For fairness, we also train separate DB and TI models, each with the single image input used for its evaluation.

We use these methods to generate output images with the \textit{20k} subject-prompt pairs in Stellar. During inference, an image matting method \cite{img_matting} is used to extract human subject masks for StellarNet. We present our primary findings and analyses in the following subsections. Additional experiments and detailed results are in Supp.

\subsection{Validating the Proposed Metrics}\label{sec:exp-metrics}
    We quantitatively compare our proposed metrics with metrics typically used in existing T2I studies (Aesthetic Score~\cite{LAION-5B}, CLIP$_I$~\cite{clip_metric}, DreamSim~\cite{fu2023dreamsim}, CLIP$_T$~\cite{clip_metric}, HPSv1~\cite{wu2023better}, HPSv2~\cite{wu2023human}, ImageReward~\cite{xu2023imagereward} and PickScore~\cite{PickaPic}) over the entire Stellar Dataset.

    \paragraph{Alignment with Human Judgment.}
        To assess the alignment of ours and pre-existing metrics with human judgment, we perform two studies collecting a total of \textit{2.5k} responses using AMT. In the quantitative analysis involving these responses, we only consider answers where there is majority-agreement among the evaluators (2 out of 3 votes being the same), which account for $95\%$ of all responses. We present the results of these studies in Fig. \ref{fig:amt-study}.

        The first study, \textit{Overall}, presents the human evaluators with the output image of each method conditioned on the same subject and prompt and asks them to choose the one they prefer the most considering both the alignment of the output with the input human subject and the prompt. Our second study presents the evaluators with two images generated from two systems on the same conditioning (we pair methods at random in each trial), ignoring the representation accuracy of the input subject, and asks them to \Circled{\textbf{1}} select the one whose objects in the output image best align with the prompt, \textit{Obj}, and \Circled{\textbf{2}} select the one that better aligns the human-objects relations with the prompt, \textit{Rel}.

        We start by computing the Kendall-$\tau$ rank correlation between the collected human responses and different metrics. We find our metrics complementing each other with \texttt{IPS} outperforming existing metrics in human preference alignment. As expected, \texttt{GOA} and \texttt{RFS} excel in the \textit{Obj} and \textit{Rel} studies, respectively, indicating the efficacy of our proposed object-centric context metrics in decoupling and evaluating specific localized aspects of text-to-image alignment.

    \paragraph{Correlation with Existing Metrics.}
        Expanding beyond human judgment, we study the intra and inter-correlation for our metrics and pre-existing ones, revealing a low Pearson correlation coefficient of $\rho=0.180$ with existing metrics, which is \textit{3x} lower than the intra-correlation observed among established metrics ($\rho=0.471$). This analysis suggests that our metrics add a new dimension to the evaluation of T2I personalization models.
        
        Furthermore, our metrics are distinct in that they encompass various aspects of the input. For instance, object-oriented metrics like \texttt{GOA} and \texttt{RFS} exhibit a low correlation with image-based metrics (\texttt{APS}, \texttt{IPS}, \texttt{SIS}), evidenced by a Pearson correlation coefficient of $\rho=0.08$.

    \paragraph{On the Complementarity of \texttt{APS} and \texttt{IPS}.}
         \texttt{APS} and \texttt{IPS} are expected to be highly correlated. However, Fig. \ref{fig:method-evaluation} (bottom row) highlights the importance of distinguishing coarse facial features (IPS) from fine-grained facial attributes (APS). Unlike \texttt{IPS}, which struggles to accurately discern StellarNet's improved capture of the input identity due to changes in non-facial characteristics (\eg~removing the input's eyeglasses); the focus of \texttt{APS} to solely invariant face attributes manages to effectively showcase our method's enhanced ability to represent this identity.

\subsection{Surveying Personalized Generators}
    \begin{figure*}[!ht]
        \centering
        \includegraphics[width=0.95\textwidth]{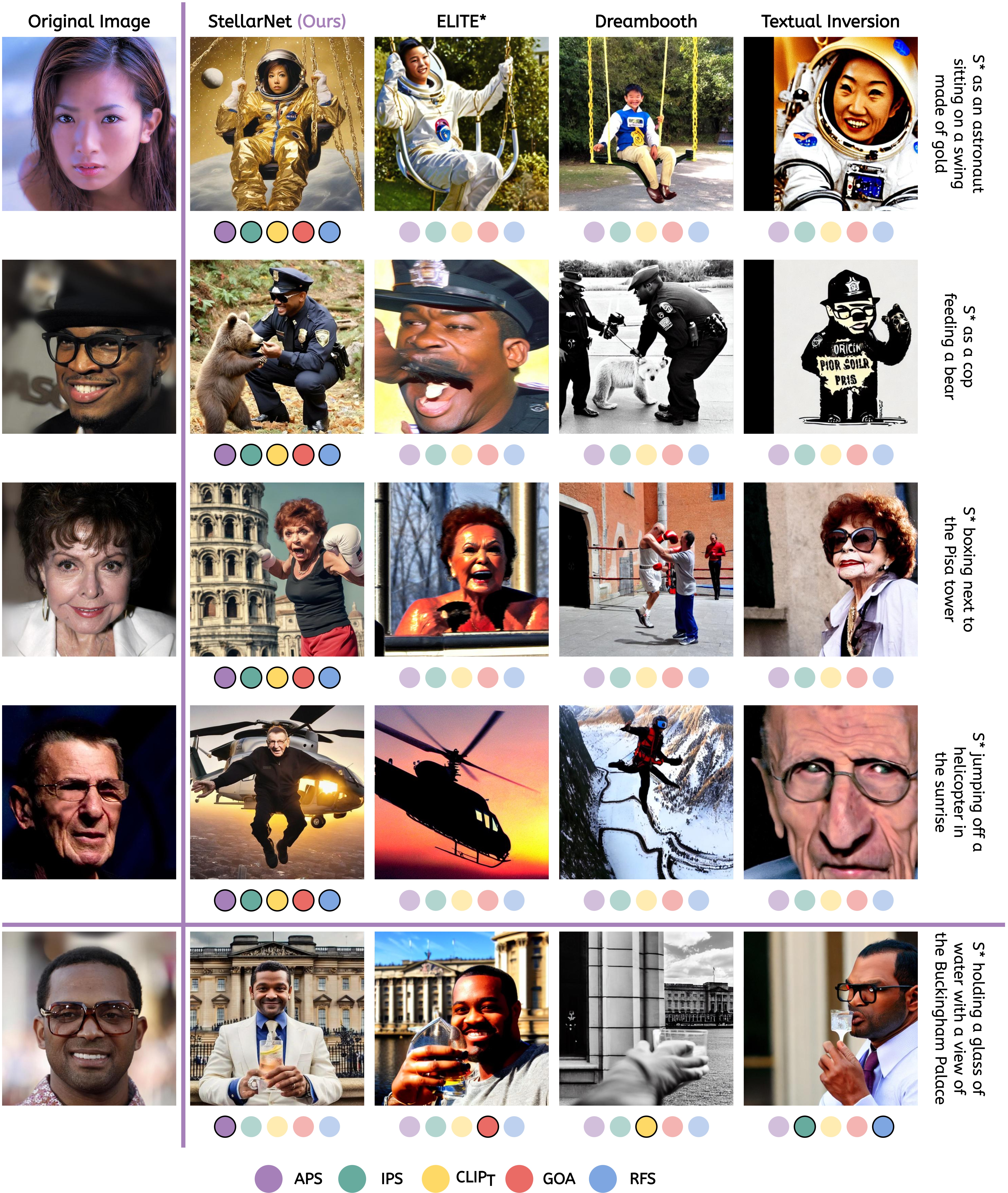}
        \caption{\textbf{Qualitative comparison of StellarNet vs. SoTA personalized-T2I methods.} The leftmost column depicts the input image from CelebAMask-HQ~\cite{CelebAMask-HQ} portraying the actor's identity (marked in text as $\mathbf{S^*}$). The four rightmost images are generations based on the system delineated in the column's title. All methods input the corresponding prompt shown next to each row. Additionally, with every generation, we include five colored circles representing the preference of the five metrics listed at the bottom of the figure. An opaque circle (\eg~\protect\coloredcircle{violet!60}), indicates the image with the highest score for each metric among the generations of the same row.}
        \label{fig:method-evaluation}
    \end{figure*}

    We compare our baseline method, \cref{sec:method}, against established personalized T2I generators. This comparison utilizes both pre-existing metrics from the literature and our newly introduced metrics from \cref{sec:metrics}, with results presented in Tab. \ref{table:quant_stellar_vs_sota}. Notably, StellarNet emerges as the top performer by a wide margin in almost all metrics. The exception is CLIP$_T$, where our performance does not surpass other methods. This aligns with findings from previous studies \cite{sdxl,PickaPic}, which also report a disconnect between CLIP$_T$ scores and human preferences. This discrepancy, as further discussed in \cref{sec:related_works}, highlights an inherent limitation of CLIP$_T$.

    Using the results from our first study, \textit{Overall} (\cref{sec:exp-metrics}) we find StellarNet's output generations to be preferred by humans when given images generated by all ablated systems (Fig. \ref{fig:amt-study} - left). StellarNet outperforms other methods by a wide margin -- it is preferred in $\mathbf{78.1\%}$ of all trials,  compared to $8.7$\% for the next best method. Additionally, Fig. \ref{fig:method-evaluation} offers a qualitative assessment of our method as compared to other methods showing the effectiveness of Stellar (2nd column). Finally, we show that our method produces diverse outputs for different random seeds while being faithful to the input identity, as shown in Fig. \ref{fig:seed-ablation}.

\section{Conclusion \& Ethical Considerations}
\label{sec:conclusions}
In this paper, we identified two impediments in the progress of personalized, human-centric T2I methods. First, there is a lack of large-scale data appropriate for this task, and second, there is a literature gap regarding existing evaluation metrics. To aid future developments, we curated Stellar: a bi-modal dataset with thousands of diverse prompts and semantically rich annotations. We also introduced an array of intuitive metrics specializing in personalization systems, which correlate significantly better with human judgment than previously used alternatives. The combination of annotated data and our explainable metrics allowed us to rigorously compare recent methods, including our simple yet powerful method and last contribution: StellarNet, which established quantitative and qualitatively a new SoTA.

\paragraph{Ethical Considerations.}
    Image generation models can be \textit{misused} as tools for generating false content or prompting misinformation from bad actors. Personalized Generative AI could be misused to create fake images of specific individuals, portraying them offensively or inappropriately. StellarNet heavily relies on SDXL, which is trained with millions of pictures from around the web. Therefore, our model inherits typical SDXL's shortcomings, such as the potential to generate biased content, exaggerating social stereotypes, or making otherwise NSFW material. One can use modern tools, such as NSFW classifiers~\cite{nsfw_function} to attempt to control and reduce inappropriate content generation (a practice we followed in this work). Nevertheless, similar to other Generative AI models, StellarNet should not be directly used in real-world applications without a careful inspection of the model's output. We urge future users to utilize our work \textit{responsibly}~\cite{ethical_use}. Moreover, we advise proper content moderation and relevant regulation~\cite{biden_exec} to prevent malpractice.




{
    \small
    \bibliographystyle{ieeenat_fullname}
    \bibliography{references}
}

\end{document}